\documentclass{article}

\usepackage[preprint,nonatbib]{nips_2018}

\usepackage[utf8]{inputenc} 

\usepackage[T1]{fontenc}    
\usepackage{hyperref}       
\usepackage{url}            
\usepackage{booktabs}       
\usepackage{amsfonts}       
\usepackage{nicefrac}       
\usepackage{microtype}      
\usepackage{amsmath}
\usepackage{amsthm, amssymb, graphicx, tikz, hyperref, ifthen, xspace, bm, graphics}
\usepackage{fancyhdr}
\newcommand{\comment}[1]{}
\usepackage[square,sort,comma,numbers]{natbib}

\usepackage{fancyhdr}

\usepackage[nameinlink,noabbrev]{cleveref}
\newcommand*{\fullref}[1]{\hyperref[{#1}]{\Cref*{#1} -- \nameref*{#1}}}

\usepackage{enumerate}

\theoremstyle{plain}

\theoremstyle{definition}

\theoremstyle{remark}

\newcommand{\R}{\mathbb{R}}

\newcommand{\set}[1]{\left\lbrace #1 \right\rbrace}

\newcommand{\abs}[1]{\left\vert #1 \right\vert}
\newcommand{\norm}[2][]{{\left\Vert #2 \right\Vert}_{#1}}

\newcommand{\vect}[1]{\bm{#1}}

\newcommand{\one}{1}
\newcommand{\indicator}[1]{\one_{\left\lbrace #1\right\rbrace}}
\DeclareRobustCommand\1{\futurelet\oneNext\oneCheck}%
\def\oneCheck{%
	\ifx\bgroup\oneNext \expandafter\indicator%
	\else%
	\expandafter\one%
	\fi%
}

\title{Improving Visual Recognition using Ambient Sound for Supervision}

\author{
  Rohan Mahadev \\
  Department of Computer Science \\
  New York University \\
  New York, NY \\
  \texttt{rm5310@cs.nyu.edu} \\
  \And
  Hongyu Lu  \\
  Department of Mathematics \\
  New York University \\
  New York, NY \\
  \texttt{hl1535@nyu.edu} \\
}

\begin{document}

\maketitle

\begin{abstract}
Our brains combine vision and hearing to create a more elaborate interpretation of the world. When the visual input is insufficient, a rich panoply of sounds can be used to describe our surroundings. Since more than $1,000$ hours of videos are uploaded to the internet everyday, it is arduous, if not impossible, to manually annotate these videos. Therefore, incorporating audio along with visual data without annotations is crucial for leveraging this explosion of data for recognizing and understanding objects and scenes. \citep{ambient} suggests that a rich representation of the physical world can be learned by using a convolutional neural network to predict sound textures associated with a given video frame. We attempt to reproduce the claims from their experiments, of which the code is not publicly available.  In addition, we propose improvements in the pretext task that result in better performance in other downstream computer vision tasks.

\vspace{3mm}
\textbf{Keywords}: self-supervision, multi-modal learning, computer vision
\end{abstract}

\section{Introduction}
When there is movement, there is sound. One of the most important functions of the midbrain is the integration between visual and auditory signals through the corpora quadrigemina. In particular, the upper layers of the superior colliculi receive information from the retina and the central nuclei of inferior colliculi receive signals from multiple auditory receptors. Then, the lower layers of the superior colliculi and the central nuclei of inferior colliculi create an audiovisual loop, where the information is learned together dynamically. \citep{scic} suggests that by incorporating visual and auditory information, we learn a more accurate perception of the world. As a result, it is natural to question the sufficiency of unimodal computer vision algorithms in interpreting the complex world we experience. 
 
 On the other hand, since the invention of the convolutional neural network \citep{cnn} and other deep learning techniques, we are able to create highly accurate supervised models that perform many computer vision tasks exceptionally well. However, training these models can be very expensive since labeling data requires exceedingly large amount of human effort and time. Alternatively, self-supervised learning has gained popularity over the past few years. It is cost-effective and it can be generalized to a multitude of downstream tasks.     

In this paper, we present a self-supervised model that learns representations of the world using statistical summaries of sound textures extracted from audio associated with the frames in videos. We then evaluate the model performance by fine-tuning the model for object classification and compare the results to those from the current state-of-the-art models.

\newpage
\section{Related Work}

\textbf{Sound representation}.  A sound texture is a composition of acoustic features. Many studies have been done to create different representations of sound textures such as Markov model-based clustering \citep{markov} and support vector machine \citep{SVM}. In our paper, however, we focused on representing sound textures using statistical summaries. Several works have approached sound representations and shown interesting results \citep{soundstats1, contextrecog}. We are particularly interested in the statistical model described in \citep{soundtexture}.

\textbf{Multi-modal learning}. In a multi-model learning algorithm, more than one modality is incorporated in the learning process. For instance, \citep{multimodalbolzmann} uses a deep Boltzmann machine to learn molti-modal data and \citep{imagesentencemapping} uses images and natural language data to create bidirectional retrieval mappings between them. A detailed list of multi-modal works can be found in \citep{multimodalsurvey}. We attempt to combine sound and visual data in hope to provide insights on computer vision tasks because it mimics how we perceive and learn the world.

\textbf{Self-supervised Learning}. Self-supervised learning has been able to discover natural features in the data. Many different supervisions have been proposed: \citep{ssvid} uses video tracking as supervision, \citep{ssrate} uses rate of co-occurrence in space and time as supervision. We use sound for supervision because sound provides rich information that are lacked from still images.

\section{Approach}

\subsection{Dataset}
We want to create a model which given an image, predicts the corresponding sound. For this, we use a subset of the AudioSet dataset \citep{gemmeke2017audio} which contains $10$-second video clips. We prefer to use AudioSet over other video datasets such as Kinetics \citep{kay2017kinetics} because videos in AudioSet have more local sound as opposed to background music that is counter-intuitive to our approach. We assume that the ambient sound caused by objects in the frame will help the model gain a better visual understanding as opposed to noise, such as from music videos. 

Although the AudioSet contains videos, we want to predict sound from only the given frame. This ensures that the learned image features can then be used for image recognition downstream tasks that will be discussed later in this paper. We then choose to use $3.75$-second clips of audios for their corresponding image frames as it has been shown that sound is temporally invariant in this window \citep{soundtexture}. Finally, we obtain around $55,000$ frames with their corresponding audio clips.

\subsection{Representing Sound}
\textbf{Statistical Summaries of Sound}. We follow closely the method suggested in \citep{ambient}, a modified version of \citep{soundtexture}, to compute a set of $502$-dimensional vectors of statistical summaries of the audio clips. We describe the method briefly as follows. To obtain a set of $32$ subband envelopes, a sound waveform $\vect{s}$ is convoluted with a bank $\set{\vect{f_i}}_{i=1}^{32}$ of bandpass filters 
\[
c_i(t) = \abs{(\vect{s}\ast \vect{f}_i)+i\mathcal{H}(\vect{s}\ast \vect{f}_i)}
\]
where $\mathcal{H}$ denotes the Hilbert transform, $i=\sqrt{-1}$, and $\ast$ is the convolution operator. The subband envelopes are then downsampled to $400$ Hz for computational efficiency and raised to $0.3$ power to simulate basilar membrane compression \citep{soundtexture}. Next, we compute several statistics for the subband envelopes. First, we estimate the marginal statistics, i.e. the means $\set{\mu_i}_{i=1}^{32}$ and the standard deviations $\set{\sigma_i}_{i=1}^{32}$ for each subband envelope $c_i(t)$, and normalize the standard deviations $\tilde{\sigma}_i=\sqrt{\sigma_i^2/\mu_i^2}$. Then, we compute the cross-band envelope correlations
\[
\rho_{jk} = \sum_t\frac{(c_j(t)-\mu_j)(c_k(t)-\mu_k)}{\sigma_j\sigma_k} 
\]
for all $j,k\in[1,\ldots,32]$ such that $\abs{j-k}\in\set{1,2,3,5}$. Lastly, we calculate the loudness $l=\mathrm{median}(\norm{\vect{c}(t)})$ of the subband envelopes. On the other hand, we estimate modulation filters by convoluting the subband envelopes with a bank $\set{\vect{m_j}}_{j=1}^{10}$ of bandpass filters 
\[
b_{ij} = \frac{\norm{c_i\ast m_j}^2}{N},
\]
where $N$ is the length of the modulation filters. The $b_{ij}$'s are then normalized by the same variance of the subband envelope $\tilde{b}_{ij}=\sqrt{b_{ij}/\mu_i^2}$. Lastly, we flatten the vectors and matrices and organize these statistics into the desired summary vector
\[
(\mu_1,\ldots,\mu_{32},\tilde{\sigma_1},\ldots,\tilde{\sigma}_{32},\rho_1, ,\ldots,\rho_{32},\tilde{b}_{1},\ldots, \tilde{b}_{320},l)\in\R^{502}.
\]

\textbf{Mel-Frequency Cepstral Coefficents (MFCCs)}. In addition to the statistical summaries of sound used in \citep{ambient}, we create a second set of acoustic representations using MFCCs \citep{mfcc}. MFCCs has been a standard method to represent sound parametrically since its invention. MFCCs first low-pass filter a sound waveform $\vect{s}$ at $5$ kHz and then sample at $10$ kHz. Next, $20$ triangular bandpass filters are used to compute MFCCs for cascaded frames $12.8$ ms apart with a $256$ Hamming window using discrete Fourier transform  
\[
\mathrm{MFCC}_i = \sum_{k=1}^{20} X_k\cos\left[i\left(k-\frac{1}{2}\right)\frac{\pi}{20}\right],
\]
where $i=1,\ldots, 155$ and $X_k$ is the log-energy output of the $k$th filter. For simplicity, we use  MFCC from Librosa \citep{librosa} and obtain a set of $20\times155$ matrices and flatten them into vectors of size $3,100$. 

\subsection{Predicting sound from images}

Now that we have a set of supervisory signal vectors, we treat our problem as a regression task and predict sound vectors from images themselves using a deep convolutional neural network. However, the difficulty in training such a model has been highlighted before by \textit{Owens et al} in \cite{ambient}. Hence, using a classification approach would be a much more lucrative option as it makes analysis of the model easier. So, we use the clustering approach as described in \citep{ambient}, where we perform $k$-means clustering on the entire dataset and categorize each frame and its corresponding audio into a cluster. These different clusters can then act as output classes which can be used as the supervisory signals while training the model.

We perform the same clustering on both representations of sound (Statistical summaries and MFCCs).

\begin{figure}
\centering
\includegraphics[width=80mm]{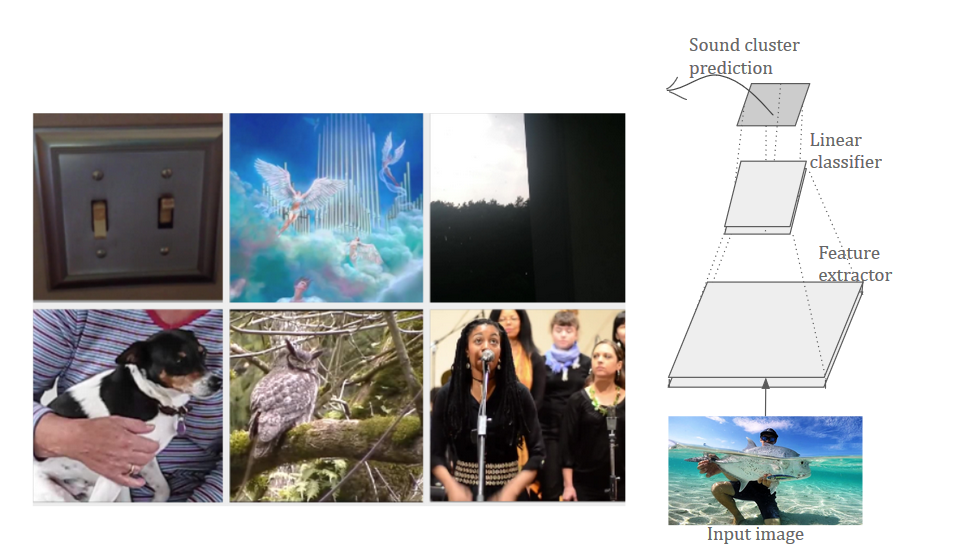}
\caption{(Left) Frames grouped by sound clusters. We see that frames from videos fall into categories such as ``outdoor scenes'', ``indoor scenes'', ``people laughing'', and ``music''. (Right) Our sound prediction model.}
\label{fig:same_cluster}
\end{figure}

The idea behind clustering frames based on their sound is that frames belonging to the same cluster would contain similar sounding objects in them. Using these clusters as our supervisory signals would help our model learn to differentiate between these clusters, and in turn, the different objects in them. These similarities can be seen in \Cref{fig:same_cluster}.

\textbf{Self-supervised training}

We use an AlexNet model \citep{alexnet}, which takes in a single RGB frame as input and predicts the corresponding cluster number (which we will call as the ``class''). We use PyTorch \citep{paszke2019pytorch} to train this model and cross-entropy loss to calculate the error between the predicted and the true class. We use the AdamW \citep{reddi2019convergence} optimizer along with the OneCycle learning rate scheduling policy \citep{smith2018super} for $200$ epochs.

An interesting question arises while training this pretext model is that to what extent should we train the model? If the model gets very good at classifying these sound clusters, it can prove degrading towards the downstream tasks. As a result, we choose to train the model for a fixed number of iterations ($200$), which usually gives us a validation set classification accuracy of nearly $45\%-55\%$ depending on the number of clusters.

\begin{figure}[ht!]
\centering
\begin{tabular}{ccccc}
\hspace{2mm} \textbf{Filter} $\vect{7}$\hspace{12mm} & \textbf{Filter} $\vect{23}$\hspace{9mm} & \textbf{Filter} $\vect{117}$\hspace{9mm} & \textbf{Filter} $\vect{129}$\hspace{9mm} & \textbf{Filter} $\vect{205}$ \vspace{2mm}
\end{tabular}

\includegraphics[width=26mm]{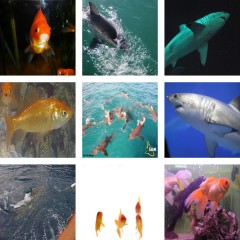}
\hspace{1mm}
\includegraphics[width=26mm]{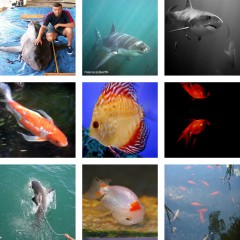}\hspace{1mm}
\includegraphics[width=26mm]{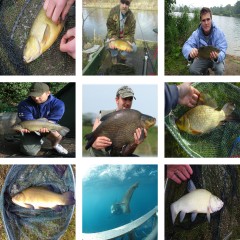}\hspace{1mm}
\includegraphics[width=26mm]{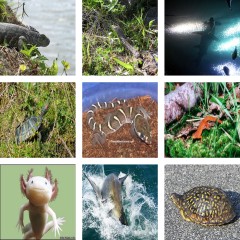}\hspace{1mm}
\includegraphics[width=26mm]{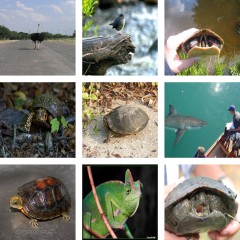}

\caption{Top activated images from a subset of the dataset for a specific target filters in the 3rd convolutional layer. We see that the images contain different objects in the same scene, like, fishes in water, people holding caught fish, or insects on the ground.}
\label{fig:max_act}
\end{figure}
\section{Evaluation of trained model}

\textbf{Qualitative evaluation: }

A good way to assess what the model has learned is to learn which input image maximizes the activation of a particular neuron in the model. This method has been described in detail in \cite{zeiler2014visualizing} and \cite{erhan2009visualizing}. \Cref{fig:max_act} shows the top $9$ activated images from a subset of our dataset. An interesting observation that emerges is that backgrounds of images in a given set are the same even if the objects in the foregrounds are different.

We suspect that this indicates the fact that backgrounds contribute more to the ambient sound. This still means that the model has learned a rich representation of the world. It can be demonstrated further by using this model to fine-tune for other downstream tasks.

\textbf{Quantitative - Finetuning on Pascal VOC: }

We evaluate our model quantitatively by fine-tuning it on image classification on Pascal VOC. The training set of Pascal VOC contains $2500$ images which makes it an ideal candidate for evaluation as it is very close to real world situations where obtaining a lot of labeled data is difficult. We follow the standard testing conditions for classification on the test set of Pascal VOC 2007 dataset and perform hyperparameter tuning on the validation set. 

We perform two different fine-tuning experiments, one where we only train the ``head'' of the model while we freeze the rest of it. And the other where we fine-tune the whole model.

\Cref{tab:all} summarizes and compares the mean average precision percentages between our approach and other self-supervised approaches. We use our $15$-cluster variant of the model for the comparisons.

\begin{table}[h!]
\centering
\renewcommand{\arraystretch}{1.2}
\resizebox{3cm}{!}{
\begin{tabular}{l|cc}
& Head & All \\\hline

Imagenet(pretrained) \citep{alexnet}
& 78.9 & 79.9 \\\hline\hline

Random \citep{krahenbuhl2015data} & 29 & 33.2 \\\hline
Pathak \textit{et al.} \citep{pathak2016context}
& 34.6 & 56.5 \\\hline
Donahue \textit{et al.} \citep{donahue2016adversarial} & 52.3 & 60.1 \\\hline
Owens \textit{et al.} \citep{ambient} & 52.3 & 61.3 \\\hline
Pathak \textit{et al.} \citep{pathak2017learning} & - & 61 \\\hline
Wang \textit{et al.} \citep{wang2015unsupervised} & 55.6 & 63.1 \\\hline
Doersch \textit{et al.} \citep{doersch2015unsupervised} & 55.1 & 65.3 \\\hline
Bojanowski \textit{et al.} \citep{bojanowski2017optimizing} & 56.7 & 65.3 \\\hline
Zhang \textit{et al.} \citep{zhang2016colorful} & 61.5 & 65.9 \\\hline
Zhang \textit{et al.} \citep{zhang2017split} & \textbf{63} & 67.1 \\\hline
Noroozi and Favaro \citep{noroozi2016unsupervised} & - & 67.6 \\\hline
Noroozi \textit{et al.} \citep{noroozi2017representation} & - & \textbf{67.7} \\\hline\hline
Our model & 52.8 & 55.1 \\
\end{tabular}   }\vspace{1mm}
\caption{Comparison of the statistical summaries Alexnet model with $15$ clusters to the state-of-the-art unsupervised approaches on Pascal VOC classification. Here, ``Head'' refers to training only the fully connected part of the model, and ``All'' refers to  fine-tuning the entire model.}
\label{tab:all}
\end{table}

\newpage
\section{Further experiments}

\textbf{ResNet vs AlexNet: }
We hypothesize that using a ResNet model, which is proven to have a higher learning capacity, would be more beneficial for this task. Hence, we follow the same procedure and evaluate the performance on the ResNet-18 variant and we compare the results with the AlexNet model. We find that the AlexNet outperforms ResNet-18, which could be attributed to the fact that the ResNet model requires more training than the AlexNet model to learn a better visual representation. However, when trained for the same amount of epochs, AlexNet seems to learn a better representation.

\begin{figure}
\centering
\includegraphics[width=55mm]{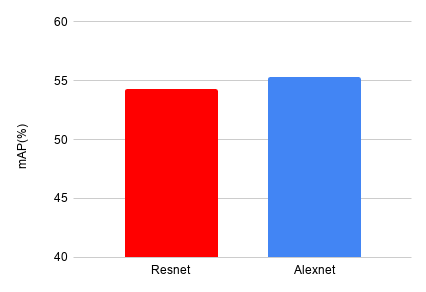}
\includegraphics[width=60mm]{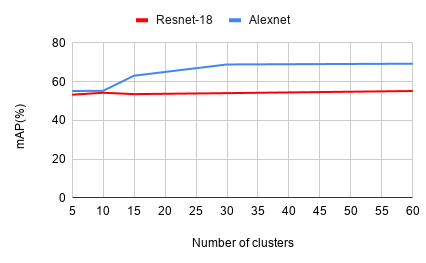}
\caption{(Left) Difference in the mean average precision obtained from two different model architectures keeping cluster size fixed at $15$. (Right) Effects of varying the cluster sizes on the two kinds of models.}
\label{fig:clusters}
\end{figure}

\textbf{Varying cluster sizes: }
We investigate how varying cluster sizes affects the performance of the model on the downstream task. As shown in \Cref{fig:clusters} We find that there is little improvement past $30$ clusters, whereas having very few clusters is also not ideal to produce good performance.

This trend is carried over when we use a ResNet model, which tells us that cluster sizes are independent of the model we use to train our pretext task.

\begin{table}[]
\centering
\renewcommand{\arraystretch}{1.2}
\resizebox{\columnwidth}{!}{
\begin{tabular}{l|cccccccccccccccccccc}
&aer & bk&brd&bt&btl&bus&car&cat&chr&cow&din&dog&hrs&mbk&prs&pot&shp&sfa&trn&tv\\\hline
Imagenet(pretrained)\citep{alexnet} & 79 & 71 & 73 & 75 & 25 & 60 & 80 & 75 & 51 & 45 & 60 & 70 & 80 & 72 & 91 & 42 & 62 & 56 & 82 & 62 \\\hline\hline
Owens \textit{et al.} \citep{ambient} & 68 & 47 & 38 & 54 & 15 & 45 & 66 & 45 & 42 & 23 & 37 & 28 & 71 & 58 & \textbf{85} & \textbf{25} & 26 & 32 & 67 & 42 \\\hline
Colorization \cite{zhang2016colorful}
& 70 & 50 & 45 & 58 & 15 & 45 & 71 & 50 & 39 & 20 & 38 & 41 & 72 & 57 & 81 & 17 & \textbf{42} & 41 & 66 & 38 \\\hline
Tracking \citep{wang2015unsupervised} & 67 & 35 & 41 & 54 & 11 & 35 & 62 & 35 & 39 & 21 & 30 & 26 & 70 & 53 & 78 & 22 & 32 & 37 & 61 & 34 \\\hline
Object motion \citep{pathak2017learning} & 65 & 39 & 39 & 50 & 13 & 33 & 61 & 36 & 39 & 24 & 35 & 28 & 69 & 49 & 82 & 14 & 19 & 34 & 56 & 31 \\\hline
Patch position \citep{doersch2015unsupervised} & 70 & 44 & 43 & 60 & 12 & 44 & 66 & \textbf{52} & 44 & 24 & 45 & 31 & 73 & 48 & 78 & 14 & 28 & 39 & 62 & 43 \\\hline
Egomotion \citep{agrawal2015learning} & 60 & 24 & 21 & 35 & 10 & 19 & 57 & 24 & 27 & 11 & 22 & 18 & 61 & 40 & 69 & 13 & 12 & 24 & 48 & 28 \\\hline
Texton-CNN \citep{texton} & 65 & 35 & 28 & 46 & 11 & 31 & 63 & 30 & 41 & 17 & 28 & 23 & 64 & 51 & 74 & 9 & 19 & 33 & 54 & 30 \\\hline
$k$-means \citep{krahenbuhl2015data} & 61 & 31 & 27 & \textbf{69} & 9 & 27 & 58 & 34 & 36 & 12 & 25 & 21 & 64 & 38 & 70 & 18 & 14 & 25 & 51 & 25 \\\hline\hline
Ours - Sound Texture & \textbf{76} & 58 & 45 & 57 & \textbf{20} & \textbf{60} & \textbf{76} & 48 & 44 & 35 & 46 & 42 & \textbf{75} & \textbf{69} & \textbf{90} & \textbf{33} & 34 & \textbf{43} & \textbf{76} & 43 \\\hline
Ours - MFCC & 74 & \textbf{60} & \textbf{48} & 57 & 20 & 54 & \textbf{76} & 49 & \textbf{45} & \textbf{37} & \textbf{51} & \textbf{43} & 74 & \textbf{69} & 87 & 31 & 40 & 42 & 75 & \textbf{45} \\

\end{tabular} }
\vspace{2mm}
\caption{Per class AP score on the Pascal VOC classification task. The average of these $20$ scores represents the mAP of the model.}
\label{tab:perclass}
\end{table}

\begin{figure}[h!]
    \centering
    \includegraphics[width=75mm]{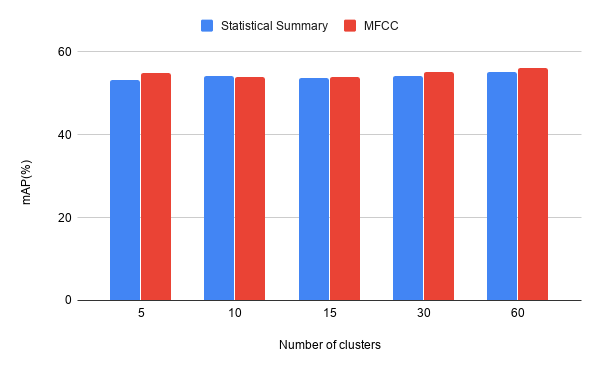}
    \caption{Comparing the mAP on the downstream task by models learned using Statistical summaries vs MFCCs as the audio representation.}
    \label{fig:feat_comp}
\end{figure}

\textbf{Statistical summaries vs MFCC: }
We choose MFCCs as an alternative sound representation. To see how the learned representation varies, we evaluate the mAP on the downstream image classification task on Pascal VOC in a similar manner using the model trained with the MFCC features. \Cref{fig:feat_comp} shows the comparison between statistical summaries and MFCC features on the downstream task. We can see that there is a slight improvement using MFCCs. This can be attributed to the different representation of the sound which leads to a different clustering, which then in turn lead to a different visual representation learned by the model.

\section{Discussion}
We see that the representation of sound matters significantly in this study. Hence, looking for an optimal sound representation will be crucial to advances in this approach. A case can be made for learning these audio representations using deep neural networks as shown in \citep{audio_cnn}. However, since they would be trained using human annotated labels, they would not be in the spirit of a truly self-supervised approach. 

The effects of using deeper networks like Wide-Resnets \citep{zagoruyko2016wide} and ResNeXts \citep{xie2017aggregated}, which currently are the state-of-the-art models for image recognition, can also be studied. However, their performance need not be analogous as seen here by the comparison between ResNet-18 and AlexNet. 

Other self-supervised techniques are evaluated on different downstream tasks such as bounding box detection and instance segmentation as well. This means that visual representations obtained in this manner are rich enough to translate to several different tasks and the limits of which can be further studied.

\section{Conclusion}
In this paper, we showed a scalable approach for training self-supervised models on image recognition tasks using audios as supervisory signals. 
We successfully reproduced the claims in \cite{ambient}. Additionally, we observed that using different representations of audios can affect the performance on the downstream task and mel-frequency cepstral coefficents tend to perform better than statistical sound summaries. Our approach works best when the pretext task is trained on a dataset like AudioSet, where the sources of sound are local to the image. This approach is highly scalable due to the abundance of available videos and therefore making it a good candidate for learning deep representations in a self-supervised manner where obtaining human annotations is proven to be difficult.

\bibliographystyle{plain}
\bibliography{reference}

\end{document}